\documentclass[conference]{IEEEtran}
\IEEEoverridecommandlockouts
\usepackage{pbalance}
\usepackage{listings}

\usepackage[dvips]{graphicx}
\usepackage{algorithm}
\usepackage[OT4,T1]{fontenc}
\usepackage[cmex10]{amsmath}
\usepackage{url}
\usepackage{multirow}

\usepackage{siunitx}
\usepackage[sort,numbers]{natbib}

\usepackage[dvips]{graphicx}

\usepackage{algorithmic} 
\usepackage{verbatim}
\title{Unused information in token probability distribution of generative LLM: improving LLM reading comprehension through calculation of expected values.}
\author{
\IEEEauthorblockN{Krystian Zawistowski}
\IEEEauthorblockA{0009-0002-9589-9030 \\ Samsung Research Poland\\
Warsaw\\
Email: k.zawistowsk@samsung.com, krystian.zawistowski@zoho.com}
}
\begin{document}

\maketitle      
\begin{abstract}  LLM text decoding  is  key component for perceived LLMs quality. We demonstrate two experiments showing that decoding methods could be improved by manipulation of token probabilities. First, we test few LLM on SummEval summary scoring dataset, to measure reading comprehension. We compare scores from greedy decoding to expected values over the next token distribution\footnote{Source code released here: \url{https://github.com/kzawisto/unused_information_llm}}. We scale logits by large temperature to increase the entropy of scores. This allows strong improvement of performance on SummEval (in terms of correlations to human judgement).  We see improvement from 6-8\% to 13-28\% for 7B Mistral and from 20\%-46\% to 37\%-56\% for Mixtral, beating GPT 4 0314 result on two metrics. Part of the gain seems related to positional bias. Secondly, we use probability-based tree sampling algorithm, to examine all most probable generations for given prompt.\end{abstract}
%

\begin{figure*}
\centering
 \includegraphics[width=400px]{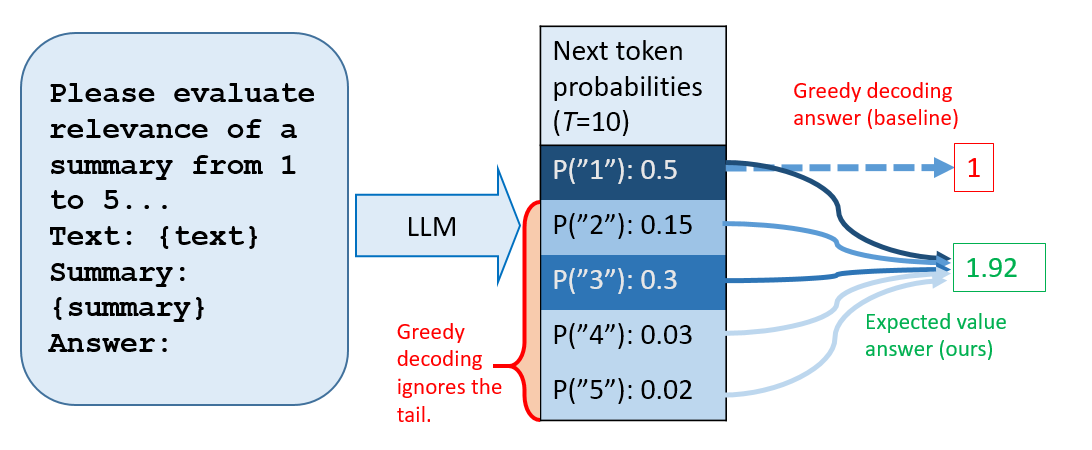}
 \caption{Conceptual diagram of presented approach: instead of answering with most probable token, we calculate  expected value for temperature $T=10$ to utilize residual information in next-token distribution.}
 \label{diag}
\end{figure*}

\section{Introduction}
\IEEEPARstart{G}{enerative LLMs} are trained on large text corpora as estimators of next token probability conditional on prior text. Then sampling from such probability distribution is performed, or token with largest probability is chosen (greedy decoding). Typically, one introduces parameter $T$ – called temperature. Let  $l_i$ be the logit for $i$-th token. Then token probability is as follows:
\begin{equation}
\label{eqproba}
 p_i =\frac{e^{\frac{l_i}{T}}}{\sum_{j}e^{\frac{l_j}{T}}}
\end{equation}

Question arises: what the T should be? Typically $T \in (0,1]$, with greedy decoding as limit in 0 and larger values corresponding to greater diversity (but also greater randomness). Research \cite{Holtzman} shows that human generated text often does not correspond to modelled highest probability. Human choice of words is not guided by greatest probability, as “humans optimize against stating the obvious”. The author of this observation, H. P. Grice in \cite{grice1975logic} gave following example: suppose that I meet a man with a gas tank asking me to sell some gas and I answer ``There's gas station over the corner''. I said only a bit of information that doesn't asnwer directly, while there's lots of implicit information not being said (gas station is open, it seels gasoline and has it available and you can go there buy some). Such concepts might be pretty foreign to LLMs: when we asked Mixtral Instruct ``How to get gasoline in Fresno'', it gave us long instruction on finding gas station on a map, choosing best gas station, operating the pump safely, paying and so on, despite the fact that most of it is irrelevant to the problem of getting gasoline in Fresno specifically.

Thus, a tradeoff arises. Probability maximization with small temperature doesn't give us natural, relevant responses. Large $T$ is not ideal either, introducing more randomness, as low probability token might be either very informative or very wrong. Is however a single fixed value of $T$ sufficient, even for specific use case?

We conjecture that decoding should be more dynamically controlled to more fully utilize the information in the distribution.

\section{Summary evaluation with expected value decoding.}
\subsection{Expected value decoding.}
Currently, greedy decoding is often used for label based QA\footnote{For example in LM Evaluation Harness, standard set of LLM benchmarks \cite{eval-harness}}. We want to find whether relative probabilities of few most probable tokens are informative. We test whether greedy decoding can be outperformed by calculation of expected value. We evaluate our approach on SummEval \cite{fabbri-etal-2021-summeval} dataset. It contains 1600 article summaries with human annotations for relevance, fluency, coherence and consistency of a summary. We compare our result against known LLM-based evaluations (\cite{wang-etal-2023-chatgpt}, \cite{shen-etal-2023-large}). LLM is asked to evaluate relevance (or other feature) on Likert scale (from 1 to 5). We use MCQ prompts from \cite{shen-etal-2023-large}, where LLM answer is A, B, C, D or E (A is 1 – worst, E is 5 – best). Let $p(A), p(B)...$ be probability of token ``A'', ``B''... Expected value score (that we use) is calculated as follows:
\begin{equation}
\label{es}
 E(s) = p(A)  + 2 p(B) + 3 p(C)  + 4 p(D)  +5 p(E),
\end{equation}
while greedy score (a standard baseline) is:
\begin{equation}
\label{smax}
 s_{max} = F\left( \textnormal{arg max}_{t \in \{A, B, C, D, E\}} p(t) \right),
\end{equation}
where $F$ is mapping $\{A\rightarrow 1, B \rightarrow 2...  \}$. $E(s)$ depends on $T$, as \eqref{eqproba} shows, while $s_{max}$ does not. However, our initial experiments show that for $T \in (0, 1]$ these values are very close to each other, $E(s) \approx s_{max}$.  LLM is spuriously certain about its answer and assigns near 100\% probability to selected answer. To fix that we increased entropy of score distribution by setting very high $T=10$ (i.e. we want for these scores to have smooth continuous distribution over the [1, 5] interval). Conceptual diagram is presented on Figure \ref{diag}.

\subsection{Results}
We evaluated Pearson correlation of scores to human judgements, the scores being calculate either with greedy method \eqref{smax} or expected value method \ref{es} with $T=10$.
We saw strong improvement in metric correlation to human judgement. In addition,
$E(s)$ scores from Mixtral 8x7B Instruct \cite{jiang2024mixtral} beat GPT3.5 and nearly match GPT4 results from \cite{shen-etal-2023-large}, see Table \ref{tab:gpt}.

\begin{table*}[]
\centering

\caption{Pearson correlations to human  judgement on
SummEval: Mixtral-Instruct and prior results for OpenAI models.}
\begin{tabular}{lccccc}
\hline
Metric & GPT3.5 0301 \cite{shen-etal-2023-large} & GPT4 0314 \cite{shen-etal-2023-large} &
ChatGPT \cite{wang-etal-2023-chatgpt} & Mixtral E(s), T=10 & Mixtral (greedy) \\
\hline
Fluency & 0.431 & \textbf{0.6} & 0.384 & 0.392 & 0.222 \\
Relev. & 0.395 & 0.461 & 0.459 & \textbf{0.555} & 0.457 \\
Consist. & 0.484 & \textbf{0.618} & 0.516 & 0.506 & 0.397 \\
Coher. & 0.416 & 0.428 & 0.438 & \textbf{0.485} & 0.427 \\
\hline

\end{tabular}
\label{tab:gpt}
\end{table*}
\begin{table*}[]
\centering
\caption{Pearson correlations to human  judgement on
SummEval for Mixtral 8x7B v 0.1 Instruct.}
\begin{tabular}{lcccc}
\hline
Metric & Fp16, greedy & Fp16, E(s) & Int4, greedy & Int4, E(S) \\
\hline
Fluency  & 0.222 & 0.392 & 0.235 & \textbf{0.405} \\
Relev. & 0.457 & 0.555 & 0.464 & \textbf{0.564} \\
Consist. & 0.379 & \textbf{0.506} & 0.293 & 0.47 \\
Coher. &0.428 & \textbf{0.485} & 0.388 & 0.438 \\
\hline
\end{tabular}

\label{mixtral}
\end{table*}

Furthermore, strong improvements were produced for small and quantized LLM too. We evaluated 3 LLMs, from 7B to 47B parameters and all metrics are consistently improved.  We compare scores for quantized and float16 Mixtral Instruct in Table \ref{mixtral}. Surprisingly, quantized Mixtral performance is only slightly worse than float16 version.

Up to 4.4 times improvement is achieved for Mistral v0.2 Instruct 7B from \cite{jiang2023mistral} (from 6.4\% to 28.4\% on relevance). We show these results in Table \ref{mistral}. In Table \ref{solar} we show result for SOLAR 10.7B Instruct \cite{kim2024solar}. While authors of the model reported it to outperform much larger Mistral 7x8B on some benchmarks, we see nothing similar for SummEval. Consistently with Mistral results, for SOLAR the largest gain was observed for relevance evaluation (from 19\% to 43\%). For every model we show results for float16 inference and also for model checkpoints quantized with use of GPTQ \cite{frantar2023gptq}.

We see that summarization metrics, being relevant automated metrics for reading comprehension, strongly improve with the number of parameters. Also, quantized 4-bit  LLMs are very strong performers proportionally to their size and outperform similar size float16 models (for instance, quantized Mixtral, having 24GB in parameters and Mixture-of-Experts architecture strongly outperforms float16 SOLAR with 21 GB of parameters). For this reason quantized LLMs might be viable, cost-effective option for RAG and other similar use cases. This phenomenon is similar to emergent abilities of LLMs \cite{wei2022emergent} where larger sizes lead to qualitative improvement in LLM performance. In this case
too, quantized LLMs \cite{liu2023emergent} retain large portion of their emergent capabilities.

Gains are particularly strong for relevance and consistency: this is important for systems that rely on reading comprehension, like RAG expert systems. We used vLLM \cite{kwon2023efficient} and Transformers \cite{wolf} for implementation.
\begin{table*}[]
\centering
\caption{Pearson correlations to human  judgement on
SummEval for Mistral 7B v 0.2 Instruct.}
\begin{tabular}{lcccc}
\hline
Metric & Fp16, greedy & Fp16, E(s) & Int4, greedy & Int4, E(S) \\
\hline
Fluency & 0.06 & \textbf{0.134} & -0.045 & 0.061 \\
Relev. & 0.064 & \textbf{0.284} & 0.074 & 0.264 \\
Consist. & 0.061 & \textbf{0.252} & 0.076 & 0.249 \\
Coher. & 0.084 & \textbf{0.199} & 0.042 & 0.176 \\
\hline
\end{tabular}

\label{mistral}
\end{table*}

\begin{table*}[]
\centering
\caption{Pearson correlations to human  judgement on
SummEval for SOLAR 10.7B Instruct.}
\begin{tabular}{lcccc}
\hline
Metric & Fp16, greedy & Fp16, E(s) & Int4, greedy & Int4, E(S) \\
\hline
Fluency & 0.187 & 0.24 & 0.187 & \textbf{0.251} \\
Relev. & 0.192 & \textbf{0.427} & 0.165 & 0.364 \\
Consist. & 0.298 & 0.331 & 0.156 & 0,194 \\
Coher. & 0.305 & \textbf{0.362} & 0.2 & 0.267 \\
\hline
\end{tabular}

\label{solar}
\end{table*}

\subsection{Positional bias.}
Previously it was reported that LLM preference for candidate responses might be altered \cite{wang2023large} by simply reordering the responses in the prompt. This effect is called
positional bias. Our experimental setup might be affected by it, as we use multi choice question answering prompts from \cite{shen-etal-2023-large}.

We modified our approach as follows: we evaluate our score for two nearly identical MCQ prompts that differ by the order of answer candidates. One prompt has answer candidates in A, B, C... order, the other in E, D, C...
order. Having done that, we average the scores for two prompts, doing that for every example we evaluate.

We performed this experiment for Mistral 7B for relevance evaluation and results can
be found in Table \ref{mistralpos}. Quite interestingly averaging out positional bias produces strong improvement for greedy decoding, while there's no big difference for $E(s)$ decoding.
Furthermore stronger improvement is found for Fp16 model, than for Int4 model. Table \ref{mistraldpopos} shows few more experiments for Nous Hermes DPO Mistral 7B model, where 
prompt candidates are put either in ascending (A, B, C... order), reversed (E, D, C... order) or random order. NaN correlations indicate that the model had predicted identical result for all test examples.  
One new result here is that random order is remarkably bad, with many correlations dropping by 30\% or more. From human point of
view prompt says exact same thing, but this is not the same for LLMs, which cannot generalize when the structure is altered. Also, 
NousHermes Mistral, undergoing more extensive finetuning and alignment, outperforms Mistral Instruct on metrics related to logical reasoning, but underperforms on fluency and coherence.

This suggests that gains from $E(s)$ method might be related to positional bias, but 
details of it are not clear without further research.

This looks related to spurious certainity of LLM we already mentioned, our conjecture
of temperature misconfiguration and improved results for very high $T=10$.
LLM, when having no good candidate hypothesis, seems to overreact to weak signals - instance
of this problem is positional bias. While averaging out provides specific solution to positional bias, setting large temperature provides general solution: as LLM might take into account more candidate hypotheses, which presently are dominated by overreaction to spurious signal.

These problems could be related to the use of softmax function in attention heads. Neural net limitations with respect to softmax and the rank of matrix under it were brought to attention by \cite{yang2018breaking} (which proposes high rank RNN). Similar problems might reemerge in case of transformers and attention, which use relatively small matrices for attention heads.
For softmax it does not matter whether signal is weak or strong, only whether it is the strongest among provided candidates. It is also true however, that weak signal supression can be learned by the attention head in the pretraining process, especially when bias matrices are added to $Q$ and $K$ -- so it is impossible to tell more without further study.

\begin{table*}[]
\centering
\caption{Pearson correlations to human  judgement on
SummEval for Mistral 7B v 0.2 Instruct - positional bias analysis.}
\begin{tabular}{lcccc}
\hline
Metric & Fp16, greedy & Fp16, E(s) & Int4, greedy & Int4, E(S) \\
\hline
Relevance - standard. & 0.064 & 0.284 & 0,074 & 0.264 \\
Relevance - average. & 0.245 & 0.295 & 0.161 & 0.307 \\
\hline
\end{tabular}
\label{mistralpos}
\end{table*}

\begin{table*}[]
\centering
\caption{Pearson correlations to human  judgement on
SummEval for Mistral 7B Nous Hermes DPO fp16 - positional bias analysis.}
\begin{tabular}{lcccccc}
\hline
Metric & Greedy, ascending & Greedy, reverse & Greedy, random & E(S), ascending  & E(S), reversed  & E(S), random  \\
\hline
Relevance & 0.32 & 0.36 & 0,092 & 0.44 & 0.42 & 0.2 \\
Consistency & 0.38 & 0.33 & 0.17 & 0.5 & 0.42 & 0.3 \\
Fluency & 0.097 & NaN & 0 & 0.12 & 0.0 & 0.015 \\
Coherence & 0.16 & NaN & 0.04 & 0.24 & 0.06 & 0.061 \\
\hline
\end{tabular}
\label{mistraldpopos}
\end{table*}

\subsection{Statistical analysis.}

We evaluate our results on 1600 samples from SummEval dataset, calculating Pearson correlation to
human judgement evaluations on four metrics: fluency, relevance, consistency, coherence.
We recalculate correlations for 1600 ChatGPT-evaluated samples provided by \cite{wang-etal-2023-chatgpt} and 1200 samples evaluated by GPT3.5 0301 and GPT4 0314, provided by \cite{shen-etal-2023-large}.

We evaluate statistical significance with use of bootstrap method. We randomly shuffle series of human
evaluation metric $x_i$  and we do it 10000 times. For every random shuffle $\hat x_i$ and Pearson correlation
coefficient $r$ we calculate $\bar x_i = \frac{r}{\sqrt{1-r^2}} \hat x_i + x_i$. Clearly for large sample size
Pearson correlation $corr(x_i, \bar x_i) \rightarrow r$. We examine the empirical cumulative distribution  $P(\rho | r, x_i)$ of $corr(x_i, \bar x_i)$ .

With this we seek to evaluate, whether the difference of two sample correlations $r_1$ for sample 1 and $r_2$
for sample 2 is statistically significant. Significant difference of $r_1$ and $r_2$  would correspond to $r_2$ being unlikely result if real correlation for sample 2 was $r_1$: $$P(\rho<r_2 | r_1, x_i ) < 5\%$$ and $$1 - P(\rho>r_2 | r_1, x_i ) < 5\%,$$ according to  $p$-value testing methods.

Our estimates suggests that our results in Table \ref{tab:gpt} for consistency evaluation
and relevance evaluation with Mixtral are significantly better that GPT model results, as far as statistical
significance is concerned.

In addition, almost all improvements of $E(s)$ method over greedy method provide significant
difference in correlation. Only exception is consistency evaluation for SOLAR for Fp16 model in Table \ref{solar}, which is not statistically significant (the difference is 3.3\% while significance threshold
corresponds to 3.4\%).

\section{Tree-based sampling}
To further develop our hypothesis we propose an LLM inference analysis method that, for a given prompt,
seeks to find all probable completions that could be generated by nucleus sampling - to give complete, nearly deterministic picture, what LLM outputs could be for given prompt. As a foundation we use tree-search based sampling algorithm. We use priority queue mechanism, where most probable completions are evaluated first (like in Dijkstra algorithm). Tree sampling (a.k.a beam search) is broadly implemented approach\footnote{Available in popular library Transformers \cite{wolf}.} in generative language models. Recently we saw very similar algorithm to ours \cite{dejan} applied to compiler optimization (highest probability output produces superior compiler parametrization). Other controlled beam search techniques used for improved natural laguage generation can be found in \cite{Meister2021DeterminantalBS}, \cite{Vilnis2022ArithmeticSP}.

We utilize priority based tree sampling to find all possible or most probable completions for nucleus sampling. Algorithm \ref{alg:cap} shows this procedure in pseudocode.
\begin{algorithm*}
\caption{Tree-crawling topP algorithm}\label{alg:cap}
\begin{algorithmic}
\REQUIRE
\STATE $t_1...t_n$ \COMMENT{Prompt input sequence.}
\STATE $\phi(t_1..t_n) \rightarrow l_i$ \COMMENT{LLM that maps token sequence  to next-token log-probabilities}
\STATE $\hat p \in (0,1)$ \COMMENT{TopP probability threshold}
\STATE $\alpha$ \COMMENT{Minimum loglikelihood of completion.}
\STATE $StopTokens$ \COMMENT{Tokens that terminate inference, such as newline or end-of-sentence.}
 \STATE $MaxSteps$ \COMMENT{Max number of LLM evaluations}
\ENSURE
\STATE $Queue \gets [(t_1..t_n, 0)]$ \COMMENT{Priority queue ordered by second argument}
\STATE $Complete \gets []$ \COMMENT{Generated sequences terminated on stop tokens.}
\STATE $Incomplete \gets []$ \COMMENT{Generated sequences terminated on minimum logprobability $\alpha$}
\STATE $i \gets 0$
\WHILE{$Queue$ not empty and $i < MaxSteps$}
	\STATE $seq_i, lproba \gets Queue.pop()$
	\STATE $l_i \gets \phi(seq)$
	\FORALL{$token, l\in TopPCandidates(l_i,\hat p )$}
			\STATE $el\gets (concat(seq, token), lproba+l)$
			\IF{$token \in StopTokens$}
				\STATE $Complete.append(el)$
			\ELSIF{$l+x \geq \alpha$}
				\STATE $Queue.append(el)$
				\ELSE
 				\STATE $Incomplete.append(el)$
			\ENDIF
		\ENDFOR
		\STATE $i \gets i + 1$
\ENDWHILE
\STATE \RETURN $Complete, Incomplete$
\end{algorithmic}
\end{algorithm*}
This algorithm has exponential asymptotic complexity: every iteration produces $N$ new sequences without fixed lower bound for $N$ (LLM tokenizers have tens of thousands of tokens), leading to exponential divergence $K^N$ for $K$ new tokens. One could decrease $N$ by adjusting $T$ and $\hat p$. We notice that for some prompts $N$ is small number and large values of $N$ indicate a qualitative change in the text generation (such as going from direct answer to user query, to additional not needed remarks).
As an example of this we evaluate following prompt
for Mixtral instruct\footnote{The prompt contains no new lines, but line wrapping was added for clarity.}:
\begin{small}

\begin{verbatim}
<s> [INST]Please provide one original,
creative paraphrase for sentence
"My name is John Kennedy"
and write new line after it[/INST]
Answer:\n\n"
\end{verbatim}
\end{small}
Outputs with their evaluated probabilities can be found in Table \ref{jfk}. We used nucleus sampling threshold $\hat p=0.9$ and temperature $T=2$ and we show outputs with $p>0.1\%$. Temperature is large, and reason is that for smaller temperatures only first, most probable output would be generated, while now it is generated with 73\% probability.  Output distribution is uneven, with top output 41 times more probable than second most probable output and very fat tailed with with about 23\% of probability mass distributed among very unlikely outputs ($p<0.1\%$).

We did not get any diversity of the paraphrase, despite asking for it explicitly, we got only two options:
``I go by the name of Kennedy John'' and ``I go by the name of JFK''. At the same time
LLM becomes unpredictable in the tail of the distribution as various additional comments follow
after requested text. Output is thus not diverse and diversity we get provides little benefit: it might be a problem, when long unpredictable output follows the answer, evading usual
stopping mechanisms of the inference (here we stop inference on two new lines in a row.)

This type of behavior can be easily explained with reference to the content of training corpora for LLM.
Specific tasks like ``paraphrase this sentence'' or specific sentences like ``My name is John Kennedy''
are likely rare in the large internet crawl corpora. At the same time, casual conversation is more frequent, so LLM can generate diverse full sentence answers (but this is not very useful for instruction-following tasks).
\begin{table*}
\caption{Tree sampling algorithm output for Mixtral Instruct, completed senteces.}
\begin{tabular}{ll}
\hline
Probability & LLM output  \\
\hline
  0.73   &  I go by the name of Kennedy John." \\
   0.018 & I go by the name of JFK" \\
    0.005 &  I go by the name of JFK" is a creative paraphrase for "My name is John Kennedy." \\
  0.005 & I go by the name of JFK" is a creative paraphrase for the sentence "My name is John Kennedy." \\
 0.003 & I go by the name of JFK" is a creative paraphrase of "My name is John Kennedy." \\
 0.002 & I go by the name of JFK" is a possible creative paraphrase for "My name is John Kennedy." \\
 0.001 & I go by the name of JFK" is a possible paraphrase for "My name is John Kennedy." \\
 0.001 & I go by the name of JFK" followed by, "What an honor to make your acquaintance!" \\
\hline
\end{tabular}

\label{jfk}
\end{table*}

Results point to causes of few problems of LLMs that we believe to be fairly widespread in applications
based on prompting LLM and parsing their output.
\begin{enumerate}
 \item  Repetitive output – LLM output might be often identical, as there is single completion with very large probability. LLM might sometimes provide little advantage compared to retrieval based or rule based system.
\item Stopping instability – when LLM completes desired output, many different unwanted follow-up comments might be produced, disturbing rule based inference stopping mechanism.
\item Uncontrollability – when LLM is asked to do something, it can ignore instruction.
\item Hallucination - false or otherwise unwanted outputs might be produced by LLM in rare cases,
while being undetected in tests using standard decoding.
\end{enumerate}

These issues are rarely detected by commonly used accuracy-based benchmarks. There are  some generic automated metrics such as MAUVE for diversity \cite{pillutla2023mauve}, but goal of our method is to  analyze these issues in specific use cases and provide guidelines on configuration and further analysis.
Our algorithm allows to analyze influence of modified prompts and system configuration on output probabilities, allowing greater degree of reliability and objectivity in the development, as different prompts, LLMs or sampling algorithm can be compared. On more general terms it seems that decoding heuristics like TopP seem to fail our expectations, where entropy is extremely small or extremely large. Instead other approaches might be investigated:
a) Scaling $T$ for entropy extremes of next-token probability distribution.
b) Taboo sampling – tree sampling with penalty for token and substring repetition.
c) Stopping generation for large entropy spikes (as that would likely result in unwanted output).
This might allow new improvements in few important KPI for LLM, such as output diversity, controllability and safety.

Among related work we may mention Mirostat \cite{basu2021mirostat}, a sampling algorithm similar to TopP. Fixed perplexity objective allows to avoid both incoherence for large $p$ and repetition for small $p$,  while similar approach of \cite{Meister2022LocallyTS} seeks to generate text with locally constant information content. Also vLLM \cite{kwon2023efficient} implements repetition penalty to stabilize low $p$ inference. Another tradeoff however remains, as  high Shannon information makes no difference between highly informative word in human terms and unwanted token overrepresented in the training dataset (as Shannon information is defined as the inverse of probability). In addition, prioritizing largest possible probability is very useful for some use cases, such as  multi-choice QA \cite{eval-harness} or LLM for source code generation (see Fig. 7 in \cite{chen2021evaluating}) or compiler configuration \cite{dejan}. Balancing this tradeoff for humal language is a problem with no  general solution yet known to us. We thus seek to provide a tool for analyzing specific problems, such as prompting, interaction between LLM and rule based scripts or stopping LLM inference reliably.

\section{Conclusions}
We show LLM decoding method that improves performance
for answers given on quantitative scale: such as ``evaluate relevance of summary on a scale from 1 to 5''.
On SummEval summary evaluation dataset the method produces strong improvements, with open source LLM nearly matching much larger GPT3.5 and GPT4, with GPT4 0314 outperformed on relevance and coherence. Such improvement supports our hypothesis that the
temperature might be not optimally configured in standard LLM decoding, as token probabilities do not reflect real world probabilities and small and large temperatures serve different purposes.

We demonstrate new LLM analysis method using priority based tree sampling algorithm, useful for study of some niche problems in LLM, such as the diversity and controllability of the
output.

We show reading comprehension metrics for few different LLM with sizes 7B, 10.7B and 47B with float16
(half precision) inference and 4 bit GPTQ quantization.
Summarization metrics strongly improve with the number of parameters,
and quantized 4-bit  LLM are  effective in proportion to their size (which is of interest for RAG on low-powered systems).

\bibliographystyle{myIEEEtran}

\bibliography{fedcsis_version}
\end{document}